\documentclass[runningheads]{llncs}

\usepackage{eccv}

\usepackage{eccvabbrv}

\usepackage{graphicx}
\usepackage{booktabs}

\usepackage{algorithm}
\usepackage[noend]{algpseudocode}
\usepackage{caption}
\usepackage{subcaption}
\usepackage{multirow}
\usepackage{booktabs}
\usepackage{amsmath}
\usepackage{amssymb}
\usepackage{wrapfig}
\usepackage{tabularx}

\usepackage[accsupp]{axessibility}  

\usepackage{hyperref}

\usepackage{orcidlink}

\begin{document}

\title{Dataset Quantization with Active Learning based Adaptive Sampling} 

\titlerunning{Dataset Quantization with Active Learning based Adaptive Sampling}

\author{Zhenghao Zhao\orcidlink{0009-0000-1934-4661} \and
Yuzhang Shang \and
Junyi Wu \and
Yan Yan}

\authorrunning{Z.~Zhao et al.}

\institute{Illinois Institute of Technology \\
}

\maketitle

\begin{abstract}
Deep learning has made remarkable progress recently, largely due to the availability of large, well-labeled datasets. However, the training on such datasets elevates costs and computational demands. To address this, various techniques like coreset selection, dataset distillation, and dataset quantization have been explored in the literature. 
Unlike traditional techniques that depend on uniform sample distributions across different classes, our research demonstrates that maintaining performance is feasible even with uneven distributions. 
We find that for certain classes, the variation in sample quantity has a minimal impact on performance. Inspired by this observation, an intuitive idea is to 
reduce the number of samples for stable classes and increase the number of samples for sensitive classes to achieve a better performance with the same sampling ratio. Then the question arises: how can we adaptively select samples from a dataset to achieve optimal performance? In this paper, we propose a novel active learning based adaptive sampling strategy, \textbf{D}ataset \textbf{Q}uantization with Active Learning based \textbf{A}daptive \textbf{S}ampling (\textbf{DQAS}), to optimize the sample selection. In addition, we introduce a novel pipeline for dataset quantization, utilizing feature space from the final stage of dataset quantization to generate more precise dataset bins. Our comprehensive evaluations on the multiple datasets show that our approach outperforms the state-of-the-art dataset compression methods. Code will be available at \url{https://github.com/ichbill/DQAS}.
\end{abstract}

\section{Introduction}
\label{sec:intro}

Deep learning has experienced remarkable growth recently, transforming numerous fields from image recognition~\cite{dosovitskiy2020image} to natural language processing~\cite{chowdhary2020natural}. A key driver of this success is the availability of large, well-labeled datasets, which have become the cornerstone for training large and sophisticated models. However, these extensive datasets bring with them increased computational costs and resource demands. This challenge underscores the need for effective data management techniques, such as coreset selection and dataset distillation, which aim to reduce the size of datasets while preserving their utility for model training.


Coreset selection~\cite{sener2017active_cs_3} and dataset distillation~\cite{wang2018dataset} are two pivotal techniques developed to address the challenges posed by large datasets. Coreset selection involves identifying a representative subset of a dataset that can effectively encapsulate the full dataset's characteristics. This technique allows for the training of models on smaller datasets without significant loss in performance. Dataset distillation, on the other hand, compresses data into a more compact form, ensuring models can be trained more efficiently without compromising their learning ability. Both techniques are instrumental in reducing the computational load and expediting the training process. 
Compared with coreset selection and dataset distillation techniques, dataset quantization (DQ)~\cite{zhou2023datasetdq} is a recently proposed framework to effectively compress large datasets. It is a unified dataset compression method that generates compact datasets useful for training various network architectures while maintaining high performance under all data keep ratios. The technique details of dataset quantization will be discussed in Section~\ref{sec:preliminaries}.
For simplicity, we refer coreset selection methods, dataset distillation methods, and dataset quantization methods together as dataset compressing methods in the following sections.

\begin{wrapfigure}{r}{0.6\textwidth}
    \vspace{-1cm}
    \centering
    \includegraphics[width=0.58\textwidth]{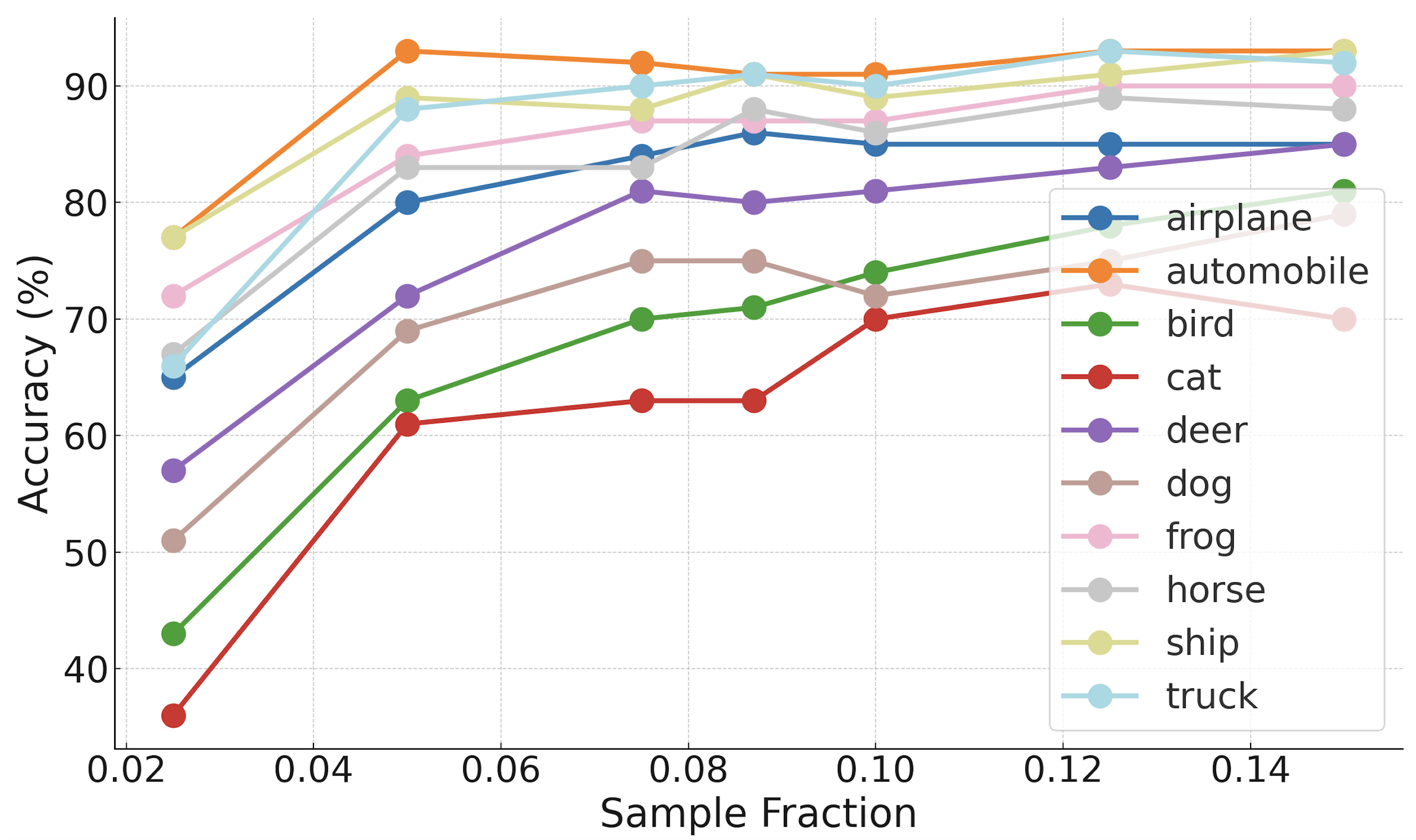}
    \caption{\textbf{Accuracy by category and sample fraction visualization on CIFAR-10.}
    This figure illustrates the outcomes of applying Dataset Quantization (DQ) across various sample fractions, followed by an evaluation of model accuracy for each category. It reveals that not all classes benefit equally from an increase in the number of samples.}
    \label{fig:obs}
    \vspace{-0.6cm}
\end{wrapfigure}

We analyze existing dataset compression methods from the perspective of their sample selection strategies.
Most existing methods use an evenly distributed subset among categories, a strategy beneficial for training and mitigating dataset bias. However, we argue that appropriate imbalanced distribution among different categories can further improve the model performance. 
On the one hand, certain classes, which we refer to as ``stable classes'', are more straightforward for the model to learn, with their images closely clustered in the feature space. In these cases, increasing the sample count does not notably improve the accuracy. Conversely, ``sensitive classes'' exhibit numerous ``outlier'' samples, where the image features in such categories are more dispersed, necessitating a larger volume of training data to achieve superior test accuracy.
For instance, we evaluate class-wise accuracy at various sampling ratios for dataset quantization, and the results are presented in Figure~\ref{fig:obs}.
For certain classes, accuracy either plateaus or increases marginally as the sample size grows. Taking the ``automobile'' class as an example, the accuracy of the model trained on a subset with a 0.05 sampling fraction is similar to that with a 0.15 sampling fraction, despite the latter having triple the number of samples. Conversely, some classes demonstrate a pronounced sensitivity to sample quantity. A notable example is the ``bird'' class, where the accuracy consistently improves with the addition of more data. These observations underscore the varying impacts of sample quantity on different classes, requiring a well-designed approach to sample a proper amount of data for each category during dataset compression.


Based on our observation, a straightforward strategy involves reducing the number of samples for stable classes and increasing them for sensitive classes. However, a key challenge lies in accurately identifying which classes are sensitive and determining the optimal number of samples to allocate to each class. 
To address this, we propose a novel approach to dataset quantization, utilizing active learning based adaptive sampling to optimize the sample selection. Active learning~\cite{sener2017active_cs_3} involves iteratively selecting the most informative and representative data samples for labeling. 
This is particularly useful in identifying how sensitive a class is, as it allows the model to focus on samples that are more challenging or have greater potential to improve the model's performance. Through active learning, we can dynamically assess and quantify the sensitivity of each class based on how the model's performance improves with the addition of new samples. In addition, to augment the efficiency of active learning's selection process, with the insight of observations in Figure~\ref{fig:obs}, we introduce a class-wise pool initialization. The class-wise pool initialization enables active learning to start from a well-estimated sample distribution across all classes.
Additionally, we introduce a novel dataset quantization architecture, which leverages the dataset features in the final stages of DQ to generate more representative dataset bins. 
This architecture is designed to maintain consistency in dataset features when dropping patches, ensuring that the dataset bin generation is unaffected.

The main contributions of this paper are summarized as follows: \textbf{(i)} We observe that the requirement for sample quantity varies significantly across different classes to optimize model training in dataset compression. \textbf{(ii)} We propose a novel dataset quantization approach, \textbf{D}ataset \textbf{Q}uantization with Active Learning based \textbf{A}daptive \textbf{S}ampling (\textbf{DQAS}). This method employs active learning with a class-wise initialization to optimize the sampling distribution. We further propose a new architecture of the dataset quantization pipeline to utilize the features of the final stage to optimize the dataset bin generation. \textbf{(iii)} We evaluate our methods on CIFAR and Tiny ImageNet datasets. The results demonstrate that our approach surpasses the performance of existing state-of-the-art dataset compression methods.

\section{Related work}
\label{sec:related_work}

\textbf{Dataset compression.} 
In this paper, we refer to coreset selection methods, dataset distillation methods, and dataset quantization methods together as dataset compressing methods in the following sections.
Dataset distillation (DD)~\cite{wang2018dataset} aims to compress a large dataset into a small synthetic dataset while preserving the performance of the models trained on it. Dataset distillation methods can be categorized into three classes: Performance matching, Parameter matching, and Distribution matching~\cite{yu2023datasetsurvey}.  Performance matching~\cite{wang2018dataset, deng2022rememberperformance_1, zhou2022datasetfrepo} aims to optimize a synthetic dataset such that neural networks trained on it could have the lowest loss on the original dataset. 
Single-step parameter matching~\cite{zhao2020datasetparameter_1, lee2022datasetparameter_2, jiang2023delvingparameter_3, kim2022datasetparameter_4, zhang2023acceleratingparameter_5, zhao2021dataset}, i.e., gradient matching, is to train the same network using synthetic dataset and original dataset for a single step, and encourage the consistency of their trained model parameters. Multi-step parameter matching~\cite{cazenavette2022datasetparameter_multi_1, li2023datasetmulti_2, cui2023scaling_multi_3, du2023minimizing_multi_4, liu2023dream, guo2023towardsdatm}, i.e., trajectory matching, the model is trained on the synthetic dataset and the original dataset for multiple steps, and minimize the distance of the endings of two trajectories. The distribution based approaches~\cite{zhao2023dataset_distribution_1, wang2022cafe_distribution_2,shang2024mim4dd} aim to obtain synthetic data, the distribution of which can match the real dataset.

Coreset selection, different from dataset distillation, seeks to identify a subset of samples that most effectively represents the entire dataset. Various criteria have been proposed for this selection process. These include geometry-based approaches as discussed in \cite{chen2012super_cs_1,agarwal2020contextual_cs_2,sener2017active_cs_3,sinha2020small_cs_4}, uncertainty-based methods \cite{coleman2019selection_cs5}, error-based techniques \cite{toneva2018empirical_cs7,paul2021deep_cs6}, and those focusing on decision boundaries \cite{ducoffe2018adversarial_cs8,margatina2021active_cs9}. Additionally, approaches like gradient-matching \cite{mirzasoleiman2020coresetscs_10,killamsetty2021grad_cs11}, bi-level optimization \cite{killamsetty2021glister_cs12}, and submodularity-based methods \cite{iyer2021submodular_cs13} have been explored.

Dataset quantization (DQ)~\cite{zhou2023datasetdq} is a recently proposed framework to effectively compress large datasets. It is a unified dataset compression method that generates compact datasets useful for training various network architectures while maintaining high performance under all data keep ratios. To address these limitations, dataset quantization (DQ)~\cite{zhou2023datasetdq}, a new framework to compress large-scale datasets into small subsets that can be used for training any neural network architectures, has been proposed. The workflow of dataset quantization is shown in Figure~\ref{fig:structure}(a). The whole pipeline is divided into three stages: dataset bin generation, bin sampling, and pixel quantization.


Despite the advancement of various dataset compression methods, the Image Per Class (IPC) count is fixed across all categories. However, as pointed out in Section~\ref{sec:intro}, an appropriately imbalanced distribution among different categories can further enhance model performance. In Section~\ref{sec:method}, we will detail our methods to adaptively sample the data.

\noindent \textbf{Active learning for sample selection.} 
Active learning aims to achieve better performance with minimal query cost by selecting informative samples from an unlabeled pool to label. Thus, active learning is mutually beneficial with coreset selection methods. For example, Sener et al.~\cite{sener2017active_cs_3} integrate the k-Center Greedy algorithm into active learning. Similarly, to identify data points close to the decision boundary, Cal~\cite{margatina2021active_cs9} employs contrastive active learning to select samples whose predictive likelihood significantly deviates from their neighbors, thereby efficiently constructing the coreset.
However, the use of active learning for coreset selection from scratch can be exceptionally time-intensive. In this paper, we introduce a novel approach: class-wise dataset initialization. This method is designed to significantly expedite the active learning process, offering a more efficient pathway to constructing effective subsets.

\section{Methodology}
\label{sec:method}

\subsection{Preliminaries}
\label{sec:preliminaries}
\textbf{Problem formulation.}
We refer to coreset selection, dataset distillation, and dataset quantization together as dataset compression. The problem of dataset compression is formulated as follows.

Assume we are given a large dataset $\mathcal{T}=\{(\mathbf{x}_i, y_i)\}^{|\mathcal{T}|}_{i=1}$ 
with $|\mathcal{T}|$ samples, where $\mathbf{x} \in \mathcal{X}$, and $y \in \{0, \dots, c-1\}$. $\mathcal{X} \subset \mathbb{R}^d$ is a $d$-dimension input set, 
and $c$ is the number of classes. We aim to learn a neural network model $\phi$ with model parameters 
$\mathbf{\theta}$ to predict labels 
$y$ for unseen images 
$\mathbf{x}$. The training object is to minimize the objective loss on the dataset, which can be formulated as:
\begin{equation}
    \mathbf{\theta}^\mathcal{T} = \underset{\theta}{\arg\min} \mathcal{L}^\mathcal{T}(\mathbf{\theta}),
\end{equation}
where $\mathcal{L}^\mathcal{T}$ is the total loss of the model $\phi_{\mathbf{\theta}}$ on dataset $\mathcal{T}$, and $\mathbf{\theta}^\mathcal{T}$ is the optimal parameter of model $\phi$ on the dataset $\mathcal{T}$.

Our goal is to generate a small dataset, $\mathcal{S}=\{(\mathbf{\hat{x}}_i, \hat{y}_i)\}^{|\mathcal{S}|}_{i=1}$, where $|\mathcal{S}| \ll |\mathcal{T}|$. For dataset $\mathcal{S}$, we can also train a model $\phi_\mathbf{\theta}$ so that
\begin{equation}
    \mathbf{\theta}^\mathcal{S} = \underset{\theta}{\arg\min} \mathcal{L}^\mathcal{S}(\mathbf{\theta}),
\end{equation}
where $\mathbf{\theta}^\mathcal{S}$ is the optimal parameter of model $\phi$ on the dataset $\mathcal{S}$. We aim to find the optimal small dataset $\mathcal{S}^*$, so that the optimal parameter of the model trained on it, $\phi_{\mathbf{\theta}^\mathcal{S}}$, minimizes the loss on the original dataset $\mathcal{T}$, which can be formulated as:
\begin{equation}
    \mathcal{S}^* = \underset{\mathcal{S}}{\arg\min}  \mathcal{L}^\mathcal{T}(\mathbf{\theta}^\mathcal{S}(\mathcal{S})),
\end{equation}
where $\mathbf{\theta}^\mathcal{S}(\mathcal{S}) = \underset{\theta}{\arg\min} \mathcal{L}^\mathcal{S}(\mathbf{\theta})$.

\noindent \textbf{Dataset quantization.}
As discussed in Section~\ref{sec:intro}, both coreset selection methods and dataset distillation methods suffer from their obvious disadvantages. On the one hand, due to the NP hard nature, coreset selection methods typically rely on heuristics criteria or greedy strategies to achieve a trade-off between efficiency and performance~\cite{yu2023datasetsurvey}. Thus, they are more prone to sub-optimal results compared with DD. On the other hand, Dataset distillation methods generate data based on certain type of task and certain network architecture, the generalization is limited~\cite{zhou2023datasetdq}. Also, dataset distillation takes a relatively long time to obtain the subset, costing a lot of time and computational resources~\cite{zhou2023datasetdq}.

To address these limitations, dataset quantization (DQ)~\cite{zhou2023datasetdq}, a new framework to compress large-scale datasets into small subsets that can be used for training any neural network architectures, has been proposed. The workflow of dataset quantization is shown in Figure~\ref{fig:structure}(a). The whole pipeline is divided into three stages: dataset bin generation, bin sampling, and pixel quantization. In dataset bin generation, a coreset selection method is used to recursively select the most representative samples from the original dataset $\mathcal{T}$ to form $N$ dataset bins $\{\mathcal{T}_1,\dots,\mathcal{T}_N\}$, which $\mathcal{T}_1\cup\dots\cup\mathcal{T}_N=\mathcal{T}$. In the bin sampling, a sampler $g(\cdot, \cdot)$ is used to select samples from each dataset bin, which is formulated as:
\begin{equation}
    \mathcal{D} = g(\mathcal{T}_1,\rho)\cup\dots\cup g(\mathcal{T}_N,\rho).
\end{equation}
In the pixel quantization stage, the selected images are divided into patches, and the patches with the lowest information will be dropped. The informative patches are used to reconstruct the image via Masked Auto-Encoder (MAE)~\cite{he2022masked} before model training.

This dataset quantization framework is able to compress the large dataset efficiently while maintaining the performance. However, the dataset quantization encounters two significant challenges. First of all, the sampling function $g(\cdot, \cdot)$ in DQ is a simple uniform sampler, which is efficient but also presents a sub-optimal performance. In addition, the samples selected from the first and second stages are able to represent the entire dataset well, but the third stage changes the features of the dataset, which leads to the inconsistency of dataset features between the first stage and the third stage.

For most dataset compression methods, the number of samples for each class is naturally equally distributed among all classes. The balanced distribution among all classes intuitively forms a robust dataset with low bias. 
However, when exploring the relationship between the number of samples per category and class-wise accuracy, we observed a peculiar trend: reducing the number of samples for certain categories does not significantly impact model performance. In contrast, decreasing the sample count for other specific categories markedly affects performance.
We attribute this phenomenon to two primary factors: First, the data within certain categories with stable performance might be similar in the feature space, implying that additional data does not contribute to the model's learned parameters during training. In contrast, for the categories where the performance of the model trained is significantly affected, the data samples are sparsely distributed across the feature space. This feature sparsity of the samples requires more data to achieve higher accuracy. 

To substantiate these hypotheses, we conducted a series of experiments. Specifically, we use DQ to compress the CIFAR-10 under various sampling ratio, and test the model performance across the categories. 
The results, as depicted in Figure~\ref{fig:obs}, reveal that even at lower sampling fractions, the model's performance in certain categories remains comparable to that achieved with higher sampling rates. This finding opens up opportunities to further reduce the sample size for specific classes without losing the performance.
From Figure~\ref{fig:obs}, we can observe that for some classes, the accuracy stops increasing or increases very slowly with the increasing samples. For example, for the ``airplane'' class, the accuracy when the fraction is 0.075 and when the fraction is 0.15 are very similar, though the previous one only takes half of the samples of the later setting. Also, some classes are sensitive to the number of samples. For example, the accuracy of ``bird'' keeps increasing when we add more data. For simplicity, we refer to the classes whose performances are barely affected by the number of samples as stable classes, and refer to those whose performance keeps increasing with the number of samples as sensitive classes.

\begin{table*}[t]
\centering
\caption{\textbf{Different sensitivities to sample quantity across categories is a common phenomenon in dataset compression.} DD method~\cite{liu2023dream} experiments among various sampling fractions. From the table, we can observe the stable classes, which maintain consistent performance regardless of the IPC allocation, and sensitive classes, which show enhanced performance with increased IPC. Stable classes are denoted in \textcolor{blue}{blue}, while sensitive classes are marked in \textcolor{red}{red}. In addition, we highlight the performances that stable classes achieve in low IPC settings, and they are competitive with those in high IPC settings. }
\begin{tabularx}{\textwidth}{c|X|X|X|X|X|X|X|X|X|X}
\toprule
IPC & \textcolor{blue}{air} & \textcolor{red}{auto} & \textcolor{blue}{bird} & \textcolor{red}{cat} & \textcolor{red}{deer} & \textcolor{blue}{dog} & \textcolor{red}{frog} & \textcolor{blue}{horse} & \textcolor{blue}{ship} & \textcolor{red}{truck} \\ \hline
15 & 70.8 & 73.2 & 50.5 & 36.2 & 57.4 & 50.4 & 72.9 & 70.4 & 73.9 & 65.0 \\
20 & 71.8 & 74.9 & 51.6 & 38.2 & 56.9 & 49.0 & 73.0 & 69.1 & 74.9 & 67.5 \\
25 & \textcolor{blue}{\textbf{75.4}} & 77.0 & 54.7 & 41.3 & 55.3 & \textcolor{blue}{\textbf{56.3}} & 77.7 & 71.2 & 74.1 & 72.6 \\
30 & 74.3 & 77.7 & \textcolor{blue}{\textbf{55.3}} & 40.0 & 63.6 & 52.3 & 82.0 & \textcolor{blue}{\textbf{73.5}} & \textcolor{blue}{\textbf{80.5}} & 73.2 \\
35 & 76.3 & 81.1 & 52.0 & 41.6 & 64.6 & 53.9 & 81.3 & 73.9 & 78.7 & 70.3 \\
40 & 75.0 & 81.9 & 53.2 & 40.5 & 67.8 & 56.5 & 80.9 & 72.3 & 79.8 & 73.4 \\
45 & 73.9 & 80.4 & 54.6 & 45.4 & 60.2 & 54.6 & 77.3 & 73.6 & 76.4 & 67.5 \\
50 & 74.5 & 82.5 & 57.0 & 45.5 & 68.4 & 58.6 & 84.2 & 73.7 & 79.0 & 76.9 \\ \bottomrule
\end{tabularx}
\label{tab:obs_1}
\vspace{-0.5cm}
\end{table*}

\subsection{Observations}
\label{subsec:obs}

In addition, we find that this phenomenon happens across many existing coreset selection and dataset distillation approaches. To illustrate, we examined a state-of-the-art dataset distillation technique, DREAM~\cite{liu2023dream}, as a case study. We listed the class-wise accuracy of DREAM under various  Image Per Class (IPC) settings in table~\ref{tab:obs_1}. 
This table reveals two distinct types of classes: stable classes, which exhibit consistent performance regardless of the IPC allocation, and sensitive classes, where performance improves with increased IPC.
Stable classes are denoted in blue, while sensitive classes are marked in red. In addition, we highlight the performances that stable classes achieved in low IPC settings, and they are competitive with those in high IPC settings. 
This insight could offer an unexplored perspective to researchers in the field of dataset compression.

These observations offer valuable insights: intuitively, we can decrease the number of samples for the stable classes, and increase the number of samples for sensitive classes.
However, identifying which classes are sensitive and determining the optimal number of samples for each class is very challenging. Fortunately, active learning, as a strategy, involves iteratively selecting the most informative and representative data samples for labeling. This is particularly useful in identifying sensitive classes, as it allows the model to focus on samples that are more challenging or have greater potential to improve the model’s performance. Therefore, we propose employing active learning to achieve adaptive sampling.

\subsection{Active Learning based Adaptive Sampling}
In Section~\ref{sec:preliminaries} we mentioned that one limitation of DQ is the sampling function. DQ uses a simple uniform sampler in the bin sampling stage. However, this sampling function is unaware of class-wise accuracy, stable classes, and sensitive classes. 
Instead, we propose an adaptive sampling method based on active learning.
Specifically, this method employs an error reduction strategy in the active learning process.
In addition, to sample the data more effectively, inspired by our observations illustrated in Section~\ref{subsec:obs}, we propose a class-wise dataset initialization. We will first introduce the class-wise dataset initialization and then introduce the procedure of the adaptive sampling.

\noindent \textbf{Class-wise dataset initialization.}
The direct use of active learning for subset sampling from scratch can be exceptionally time-intensive. 
To make the adaptive sampling procedure more efficient, we propose a class-wise initialization mechanism for the pool initialization in active learning. 


Inspired by the observation in Section~\ref{subsec:obs}, we initialize the dataset sampling based on the class-wise performance with various sampling ratios. The adaptive sampling strategy is shown in Algorithm~\ref{alg:init}.
As shown in Algorithm~\ref{alg:init}, we process an adaptive sampling for the subset initialization. 
The high-level idea is to allocate fewer samples to classes that are minimally affected by sample quantity and more samples to those significantly impacted by it.
Specifically, we initialize $\mathcal{D}$ with even distribution as a comparison standard, and then we train the model on $\mathcal{D}$.
$(\mathbf{x}_c,\textbf{y}_c)$ are the (data, label) pairs of class $c$. $\mathcal{A}=\{Acc(\phi_\mathbf{\theta}(\mathbf{x}_1),y_1),\cdots, Acc(\phi_\mathbf{\theta}(\mathbf{x}_c),y_c)\}=\{a_1,\cdots,a_c\}$ is the class-wise model performance, i.e., the prediction accuracy of the model on each class. We initialize the dataset $\mathcal{D}^0$ with a random fraction for each class. At each iteration $i$, we train the model on the dataset $\mathcal{D}^i$ and then evaluate the model on the original dataset $\mathcal{T}$. $\mathcal{A}^i$ is the class-wise model performance on iteration $i$. Then we update the $\mathcal{A}$ with $\mathcal{A}^i$. Specifically, for each class $c\in C$, we update $a_c$ with $\max(a_c,a_c^i)$.
Then we update the sampling fraction $\rho_c$ with $\mathcal{A}$ and $\mathcal{A}_i$. We update $\rho_c$ for each class $c$. If $a_c < lb$, where $lb$ is the lower bound, it means the best accuracy in iteration history is not good for class $c$, which will lead to the model performance is always low on class $c$. In this case, a random fraction is assigned to class $c$. On the other hand, if $a_c \geq lb$, we update the sampling rate for class $c$ with $\rho_c^{i+1} = \rho_c^i \times (1 + (a_c-a_c^i))$. The equation tends to increase the fraction $\rho$ if $a_c^i$ is much smaller than the achievable accuracy $a_c$. The fractions among all classes are then normalized to keep the sum of the fractions equal to 1.

\noindent \textbf{Error reduction strategy based active learning.}
As discussed in the previous section, class-wise initialization serves as a basic approach to data sampling. In this paper, active learning is employed to refine the sampling process further. The high-level concept involves augmenting the initial sample set by selectively adding samples that promise the greatest improvement, irrespective of their class affiliations. Our primary objective is to enhance evaluation accuracy on the original dataset. To achieve this, we incorporate the principle of expected error reduction~\cite{roy2001toward}. This method allows us to systematically select from candidates, thereby optimizing the overall data sampling process.

\begin{figure}[t]
\begin{algorithm}[H]
\caption{class-wise dataset initialization.}\label{alg:init}
\begin{algorithmic}[1]
\item $\mathcal{D} = g(\mathcal{T}_1,\rho)\cup\dots\cup g(\mathcal{T}_N,\rho) = \mathcal{D}_1\cup\cdots\cup\mathcal{D}_c$
\item $\mathcal{A} = \{Acc(\phi_\mathbf{\theta}(\mathbf{x}_1),y_1),\cdots, Acc(\phi_\mathbf{\theta}(\mathbf{x}_c),y_c)\} = \{a_1,\cdots,a_c\}$
\item $\rho = \{\rho_1,\cdots,\rho_c\}$, where $\rho_i \sim U[0,1]$
\item $\mathcal{D}^0 = g(\mathcal{D}_1,\rho)\cup\dots\cup g(\mathcal{D}_N,\rho) = \mathcal{D}_1^0\cup\cdots\cup\mathcal{D}_c^0$
\While{$i<max\_iter$}
    \State $\mathcal{A}^i = \{Acc(\phi_\mathbf{\theta}^i(\mathbf{x}_1),y_1),\cdots, Acc(\phi_\mathbf{\theta}^i(\mathbf{x}_c),y_c)\} = \{a_1^i,\cdots,a_c^i\}$
    \For {$c\in C$}
        \State $a_c = \max(a_c, a_c^i)$
        \If{$a_c < lb$}
            \State $\rho_c^{i+1} \sim U[0,1]$
        \Else
            \State $\rho_c^{i+1} = \rho_c^i \times (1 + (a_c-a_c^i))$
        \EndIf
        \State $\mathcal{D}_c^{i+1} = g(\mathcal{D}_c,\rho_c)$
    \EndFor
\EndWhile
\end{algorithmic}
\end{algorithm}
\vspace{-1cm}
\end{figure}

Given the original dataset $\mathcal{T}$, the active learning framework starts by assuming an unknown conditional distribution $P_0(y|\mathbf{x})$ over inputs $\mathbf{x}$ and output classes $y \in C$, alongside the marginal input distribution $P(\mathbf{x})$. $\mathcal{T}$ is then composed of (data, label) pairs sampled from $P(\mathbf{x})P_0(y|\mathbf{x})$. From these pairs, we train a model that, given an input $\mathbf{x}$, predicts the output distribution $P(y|\mathbf{x})$. Accordingly, we express the expected error of the model as follows:
\begin{equation}
    E_{P(\mathcal{T})}=\int_\mathbf{x}L(P_0(y|\mathbf{x}), P(y|\mathbf{x}))P(\mathbf{x})
\end{equation}
where $L$ is the loss function calculated by true distribution $P_0(y|\mathbf{x})$ and the model prediction, $P(y|\mathbf{x})$. 

Through class-wise dataset initialization, we get the dataset $\mathcal{D}^t$, where $t$ denotes the maximum iteration of this initialization process. We denote $\mathcal{S}$ as the current dataset in active learning, initializing it with $\mathcal{S} = \mathcal{D}^t$. Following expected error reduction~\cite{roy2001toward}, our approach is a pool-based active learning with a large candidate pool, $\mathcal{R}=\mathcal{T}\backslash\mathcal{S}$.
Within this framework, first-order Markov active learning aims to select a query, $\mathbf{x}$, that minimizes the model's error when the query's label $y$ is added to the training set, forming $\mathcal{S}\cup\{(\mathbf{x},y)\}$. 
The candidate pool $\mathcal{R}$ not only supplies a finite set of queries but also estimates $P(\mathbf{x})$.

In active learning, the true label for $\mathbf{x}$ is initially unknown before querying. The classifier provides an estimate of the distribution $P_\mathcal{S}(y|\mathbf{x})$ from which the true label of $\mathbf{x}$ is likely to be drawn. 
The estimated error for each potential label $y \in C$ is evaluated and weighted averaged by the classifier’s posterior, $P_\mathcal{S}(y|\mathbf{x})$. 
However, in our methodology, we aim to derive a subset $\mathcal{S}$ from the original dataset $\mathcal{T}$, wherein each sample pair $(\mathbf{x},y)\in\mathcal{T}$ and the true output distribution $P_0(y|\mathbf{x})$ are already established. Consequently, we bypass the need for estimating errors for each possible label $y \in C$ as in traditional active learning methods. The log loss in our approach is thus defined as:


\begin{equation}
\label{eq_1}
E_{P_{\mathbf{S}^*}} = \frac{1}{|\mathcal{R}|} \sum_{\mathbf{x}\in\mathcal{R}}\sum_{y\in C} P_0(y|\mathbf{x}) \log(P_{\mathcal{S}^*}(y|\mathbf{x}))
\end{equation}

The above loss formulation focuses on refining the model's understanding and accuracy regarding candidate samples. This approach enables the selection of examples that most effectively challenge and thus enhance the model’s existing knowledge. By incorporating these samples, the model is better equipped to understand complex patterns within the data. 

In conclusion, our active learning adaptive sampling method can be summarized in the following steps. (1) Initially, we begin by initializing the dataset $\mathcal{S}=\mathcal{D}^t$ using the aforementioned class-wise initialization approach. (2) A model is trained utilizing the current samples in $\mathcal{S}$. (3) Considering a sample from the candidate pool, $\mathcal{R} = \mathcal{T} \setminus \mathcal{S}$ as a candidate for subsequent query, and incorporating this sample $(\mathbf{x}, y)$ into the current dataset. (4) The model is retrained with this updated dataset. (5) We calculate the expected loss with Equation~\ref{eq_1}. (6) The final step involves selecting the top-K samples that exhibit the lowest expected loss, thereby optimizing our sampling process.

\begin{figure*}[t]
    \centering
    \includegraphics[width=\textwidth]{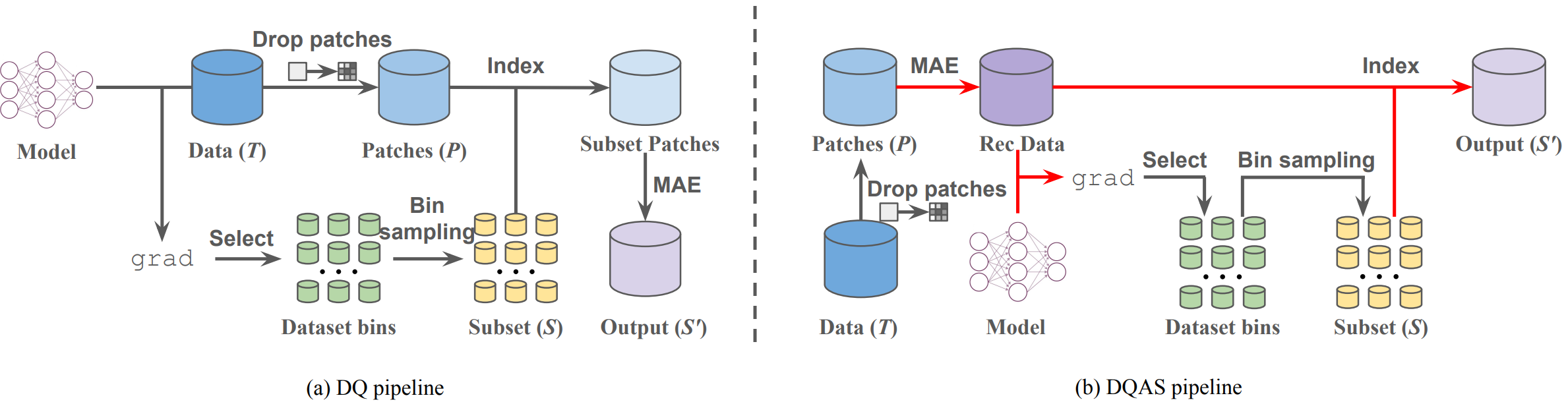}
    \caption{\textbf{Comparison between the pipeline of DQAS and DQ.} \textcolor{red}{Red} arrows are the differences between our pipeline and DQ's. The pipeline of ours
    leverages the dataset features from the reconstructed data. In this way, the dataset features remain consistent before and after dropping patches, ensuring that the dataset bin generation is not adversely affected by the patch removal and results in a more precise output $\mathcal{S}'$.
    }
    \label{fig:structure}
\end{figure*}

\subsection{Patchified-image-aware dataset quantization}
Our method is a dataset quantization method, and we propose a refined pipeline for dataset quantization. As shown in Figure~\ref{fig:structure}(a), the original dataset quantization first generates dataset bins, processes bin sampling, and drops patches with the lowest information. In the dataset bin generation, the sample is selected recursively by the GraphCut~\cite{iyer2021submodular_cs13} algorithm, and the selection of $k$-th sample in the $n$-th bin is to maximize the submodular gains $G(\mathbf{x}_k)$ as:

\begin{equation}
\label{eq_2}
    G(\mathbf{x}_k) = \sum_{p \in S^{k-1}_n} ||f(p)-f(\mathbf{x}_k)||_2^2 ~~~~~- \sum_{p \in \mathcal{T}\backslash S_1\cup\dots\cup S^{k-1}_n} ||f(p)-f(\mathbf{x}_k)||_2^2
\end{equation}

where $f(\cdot)$ is the feature extractor. However, this process suffers from the inconsistency of dataset features. 

We define the original dataset as $\mathcal{T}$, and the final output dataset as $\mathcal{S}'$. The submodular gain in the dataset bin generation, depicted in Equation~\ref{eq_2}, is computed based on the original dataset $\mathcal{T}$. However, the feature set of the output dataset $\mathcal{S}'$ undergoes dropping patches and reconstruction with MAE, leading to a discrepancy in dataset features across different stages of the workflow. This discrepancy implies that while the dataset bins are optimized for the original dataset $\mathcal{T}$, they may not be ideally suited for the derived output subset $\mathcal{S}'$, potentially impacting the efficacy of the process.

We introduce an improved pipeline for dataset quantization, as illustrated in Figure~\ref{fig:structure}(b). Since the dropping patches procedure and the image reconstruction are processed on the image level, these steps are executed initially without affecting other procedures. In a significant departure from the dataset quantization pipeline, our pipeline employs the GraphCut algorithm~\cite{iyer2021submodular_cs13} on the reconstructed dataset $\mathcal{T}'$ instead of the original dataset $\mathcal{T}$. This adjustment allows for a revision of Equation~\ref{eq_2} as follows:

\begin{equation}
\begin{split}
    G(\mathbf{x}_k) = \sum_{p \in S^{k-1}_n} ||f(p)-f(\mathbf{x}_k)||_2^2 ~~~~~- \sum_{p \in \mathcal{T}'\backslash S_1\cup\dots\cup S^{k-1}_n} ||f(p)-f(\mathbf{x}_k)||_2^2.
\end{split}
\end{equation}

This strategic adjustment ensures consistency in dataset features throughout the workflow. As a result, the dataset bins generated are optimal for the output dataset $\mathcal{S}'$, enhancing the overall effectiveness of the quantization process.

\section{Experiments}
\label{sec:experiments}

\begin{table}[t]
\centering
\caption{\textbf{Quantitative comparisons with the SOTA methods.} 
We compare our model with existing methods under various sampling ratios on multiple datasets.
The performance of our approach outperforms the existing methods.
}
\resizebox{1\linewidth}{!}{
\begin{tabular}{l|cccccccc|ccccc|ccccc}
\toprule
Dataset & \multicolumn{8}{c}{CIFAR-10} & \multicolumn{5}{c}{CIFAR-100} & \multicolumn{5}{c}{Tiny ImageNet}\\
AIPC & 50 & 250 & 500 & 1000 & 1500 & 2000 & 2500 & 3000 & 50 & 100 & 150 & 200 & 250 & 50 & 100 & 150 & 200 & 250\\
Ratio & 1 & 5 & 10 & 20 & 30 & 40 & 50 & 60 & 10 & 20 & 30 & 40 & 50 & 10 & 20 & 30 & 40 & 50\\
\midrule
Random & 36.7 & 64.5 & 75.7 & 87.1 & 90.2 & 92.1 & 93.3 & 94.0 & 52.58 & 60.48 & 65.59 & 67.61 & 71.1 & 50.19 & 52.50 & 58.52 & 61.45 & 63.83 \\
CD~\cite{agarwal2020contextual_cs_2} & 23.6 & 38.1 & 58.8 & 81.3 & 90.8 & 93.3 & 94.3 & 94.6 & 37.28 & 57.60 & 64.12 & 68.42 & 70.49 & 34.86 & 44.48 & 57.15 & 57.63 & 57.96 \\
Herding~\cite{welling2009herding_cs14} & 34.8 & 51.0 & 63.5 & 74.1 & 80.1 & 85.2 & 88.0 & 89.8 & 34.36 & 44.3 & 52.05 & 58.41 & 62.99 & 45.69 & 52.96 & 54.54 & 58.75 & 60.50 \\
k-CG~\cite{sener2017active_cs_3} & 31.1 & 51.4 & 75.8 & 87.0 & 90.9 & 92.8 & 93.9 & 94.1 & 42.78 & 59.53 & 65.61 & 68.44 & 70.3 & 46.25 & 51.30 & 59.91 & 60.78 & 63.60 \\
GraNd~\cite{paul2021deep_cs6} & 26.7 & 39.8 & 52.7 & 78.2 & 91.2 & 93.7 & 94.6 & 95.0 & 28.45 & 49.10 & 59.59 & 64.71 & 69.98 & 42.14 & 44.39 & 43.65 & 48.75 & 52.50 \\
Cal~\cite{margatina2021active_cs9} & 37.8 & 60.0 & 71.8 & 80.9 & 86.0 & 87.5 & 89.4 & 91.6 & 46.51 & 56,01 & 60.96 & 65.74 & 68.28 & 44.86 & 54.48 & 57.15 & 57.63 & 62.33 \\
DeepFool~\cite{ducoffe2018adversarial_cs8} & 27.6 & 42.6 & 60.8 & 83.0 & 90.0 & 93.1 & 94.1 & 94.8 & 46.06 & 60.54 & 65.40 & 68.12 & 70.00 & 44.29 & 49.58 & 50.79 & 58.21 & 62.94 \\
Craig~\cite{mirzasoleiman2020coresetscs_10} & 37.1 & 45.2 & 60.2 & 79.6 & 88.4 & 90.8 & 93.3 & 94.2 & 48.50 & 51.11 & 61.92 & 65.33 & 68.43 & 51.65 & 53.62 & 57.94 & 61.92 & 63.45 \\
GradMatch~\cite{killamsetty2021grad_cs11} & 30.8 & 47.2 & 61.5 & 79.9 & 87.4 & 90.4 & 92.9 & 93.2 & 42.79 & 57.85 & 64.40 & 68.72 & 69.74 & 43.23 & 46.69 & 49.10 & 51.92 & 52.41 \\
Glister~\cite{killamsetty2021glister_cs12} & 32.9 & 50.7 & 66.3 & 84.8 & 90.9 & 93.0 & 94.0 & 94.8 & 42.40 & 53.46 & 61.44 & 64.55 & 69.09 & 43.64 & 46.52 & 52.72 & 58.01 & 62.06\\
\midrule
LC~\cite{coleman2019selection_cs5} & 19.8 & 36.2 & 57.6 & 81.9 & 90.3 & 93.1 & 94.5 & 94.7 & 31.93 & 56.09 & 61.95 & 66.26 & 69.37 & 49.98 & 53.06 & 54.78 & 59.50 & 61.17 \\
Entropy~\cite{coleman2019selection_cs5} & 21.1 & 35.3 & 57.6 & 81.9 & 89.8 & 93.2 & 94.4 & 95.0 & 30.18 & 49.60 & 63.10 & 66.42 & 69.75 & 33.77 & 37.09 & 42.46 & 49.77 & 50.03 \\
Margin~\cite{coleman2019selection_cs5} & 28.2 & 43.4 & 59.9 & 81.7 & 90.9 & 93.0 & 94.3 & 94.8 & 41.51 & 59.83 & 65.08 & 68.54 & 70.54 & 37.78 & 44.58 & 44.87 & 49.88 & 53.35 \\
\midrule
FL~\cite{iyer2021submodular_cs13} & 38.9 & 60.8 & 74.7 & 85.6 & 91.4 & 93.2 & 93.9 & 94.5 & 51.43 & 60.54 & 65.96 & 68.59 & 70.38 & 44.99 & 47.62 & 49.57 & 50.00 & 56.25 \\
GC~\cite{iyer2021submodular_cs13} & 42.8 & 65.7 & 76.6 & 84.0 & 87.8 & 90.6 & 93.2 & 94.4 & 50.59 & 58.06 & 62.87 & 67.15 & 70.16 & 52.71 & 53.18 & 53.75 & 56.00 & 60.15 \\
DQ~\cite{zhou2023datasetdq} & 50.5 & \textbf{79.3} & 85.2 & 89.4 & 91.8 & 93.1 & 93.9 & 94.8 & 45.87 & 58.15 & 64.55 & 67.61 & 69.54 & 52.77 & 55.16 & 59.05 & 62.24 & 63.97 \\
Ours & \textbf{52.3} & 77.4 & \textbf{86.1} & \textbf{90.2} & \textbf{93.3} & \textbf{93.9} & \textbf{95.8} & \textbf{95.8} & \textbf{52.61} & \textbf{60.97} & \textbf{66.22} & \textbf{68.79} & \textbf{72.75} & \textbf{53.42} & \textbf{57.79} & \textbf{60.19} & \textbf{62.52} & \textbf{64.06} \\ 
\bottomrule
\end{tabular}
}
\label{tab:cifar10}
\end{table}


\subsection{Datasets and Implementation Details}
\textbf{Datasets.} Following~\cite{zhou2023datasetdq}, we mainly perform the evaluation on image classification datasets, CIFAR-10, CIFAR-100 and Tiny ImageNet. CIFAR-10 contains 50000 training images across 10 categories of common objects and 10000 images for evaluation. 
CIFAR-100 is a subset of tiny images dataset~\cite{torralba200880}, which contains 50000 samples for training and 10000 samples for testing. There are 100 classes in CIFAR-100, and each class has 500 training samples. 
Tiny ImageNet comprises 200 classes, 100,000 training images, and 10,000 test images. 

\noindent \textbf{Implementation details.}
Following the previous works~\cite{kim2022datasetparameter_4,zhou2023datasetdq}, we mainly use ResNet-18~\cite{he2016deep} in our experiments for the dataset validation. For the dataset quantization setting, we follow the dataset quantization~\cite{zhou2023datasetdq}. For experiments of dataset bin generation, we use ResNet-18 to extract features of CIFAR-10 and CIFAR-100. The models are pre-trained on the corresponding full dataset with 10 epochs. The number of bins is set to 10, and the patch dropping rate is set to 25. We use the pytorch-cifar library for model validation. We train the model for 200 epochs for CIFAR-10 and CIFAR-100 with a batch size of 128 and a cosine-annealed learning rate of 0.1. For Tiny ImageNet, we train the ResNet-50 model for 200 epochs with a batch size of 128 and a cosine-annealed learning rate of 0.6. For the active learning based adaptive sampling, we set 0.5 as the lower bound, and 50 as the maximum iteration. 

\subsection{Comparison with State-of-the-art Methods}
We compare our method to the state-of-the-art methods on CIFAR-10, CIFAR-100, and Tiny-ImageNet datasets. Specifically, we compare our methods with 14 coreset selection methods and 1 dataset quantization method. Since our method generates a dataset with imbalanced samples over classes, we use data keep ratio and Average Image Per Class (AIPC) in the experiment settings. For methods that generate balanced samples across categories, AIPC is equal to Image Per Class (IPC). For methods that generate imbalanced samples like ours, the AIPC is calculated as
\begin{equation}
    AIPC = \frac{N_{sample}}{N_{class}},
\end{equation}
where $N_{sample}$ is the total number of samples in the dataset, and $N_{class}$ is the number of classes in the dataset.

As shown in Table~\ref{tab:cifar10}, our method outperforms all existing dataset compression methods on all three datasets, which indicates the effectiveness of our method. In particular, our method outperforms the existing approaches by a large margin when the sampling ratio is low, which indicates that our methods are robust when the number of samples is small. Our method is a dataset quantization approach, while our method outperforms the previous dataset quantization method on almost all data keep ratios. Specifically, for the CIFAR-10 dataset, our method achieves a lossless result when using only 50\% of the data. 
These results indicate the robustness and efficiency of our approach.

\subsection{Ablation Study}
In this section, we perform the ablation study for the two components: active learning based adaptive sampling and patchified-image-aware dataset quantization, to prove the effectiveness of the components.


\noindent \textbf{Ablation study for active learning based adaptive sampling.}
We compare the performance metrics of our model with adaptive sampling enabled against the model where adaptive sampling is disabled. This comparison is critical in highlighting the impact of adaptive sampling on the overall efficacy of our approach. We evaluate the models on the CIFAR-10 dataset. 

The results presented in Table~\ref{tab:ablation_al} demonstrate that the inclusion of active learning significantly enhances performance compared to the approach without active learning. This observation underscores the pivotal role of active learning-based adaptive sampling in the effectiveness of our method.


\noindent \textbf{Ablation study for DQAS pipeline.}
In this section, we conduct an ablation study to evaluate the enhanced dataset quantization architecture. This involves comparing the performance of models trained on datasets created using the basic DQAS method without adaptive sampling and DQAS pipeline against those trained on datasets processed through the DQAS pipeline. The intent is to assess the efficacy of the DQAS pipeline in the dataset quantization process.
The comparative results are presented in Table~\ref{tab:ablation_arch}. A detailed analysis of these results reveals that across all ratios, the architecture developed in our approach consistently outperforms the original method. This is indicative of the robustness and efficiency of our proposed improvements in dataset quantization. The superiority of our method is particularly evident in scenarios involving complex data structures and high-dimensional spaces, where our advanced quantization technique appears to capture the underlying data distribution more effectively.

\begin{table}[t]
    \centering
    \scalebox{0.78}{
    \begin{minipage}{.65\linewidth} 
        \centering        
        \caption{\textbf{Ablation study for adaptive sampling.}}
        \begin{tabular}{c|c|c|c|c|c}
        \toprule
             Ratio & 0.05 & 0.075 & 0.1 & 0.125 & 0.2 \\
             \midrule
             DQAS w/o AS & 76.79 & 81.79 & {83.79} & {85.74} & {88.32} \\
             DQAS & \textbf{77.41} & \textbf{83.53} & \textbf{86.11} & \textbf{88.49} & \textbf{90.22} \\
        \bottomrule
        \end{tabular}  
        \label{tab:ablation_al}
    \end{minipage}%
    \begin{minipage}{.65\linewidth} 
        \centering
        \caption{\textbf{Ablation study for DQAS pipeline.}}
        \begin{tabular}{c|c|c|c|c|c}
        \toprule
             Ratio & 0.05 & 0.075 & 0.1 & 0.125 & 0.2 \\
             \midrule
             DQAS Basic & \textbf{78.47} & \textbf{81.87} & 82.94 & 85.36 & 87.83 \\
             DQ w/o AS & 76.79 & 81.79 & \textbf{83.79} & \textbf{85.74} & \textbf{88.32} \\
        \bottomrule
        \end{tabular}    
        \label{tab:ablation_arch}
    \end{minipage}
    }
    \vspace{-0.6cm}
\end{table}




\begin{wrapfigure}{r}{0.6\textwidth}
    \vspace{-1cm}
    \centering
    \includegraphics[width=0.58\textwidth]{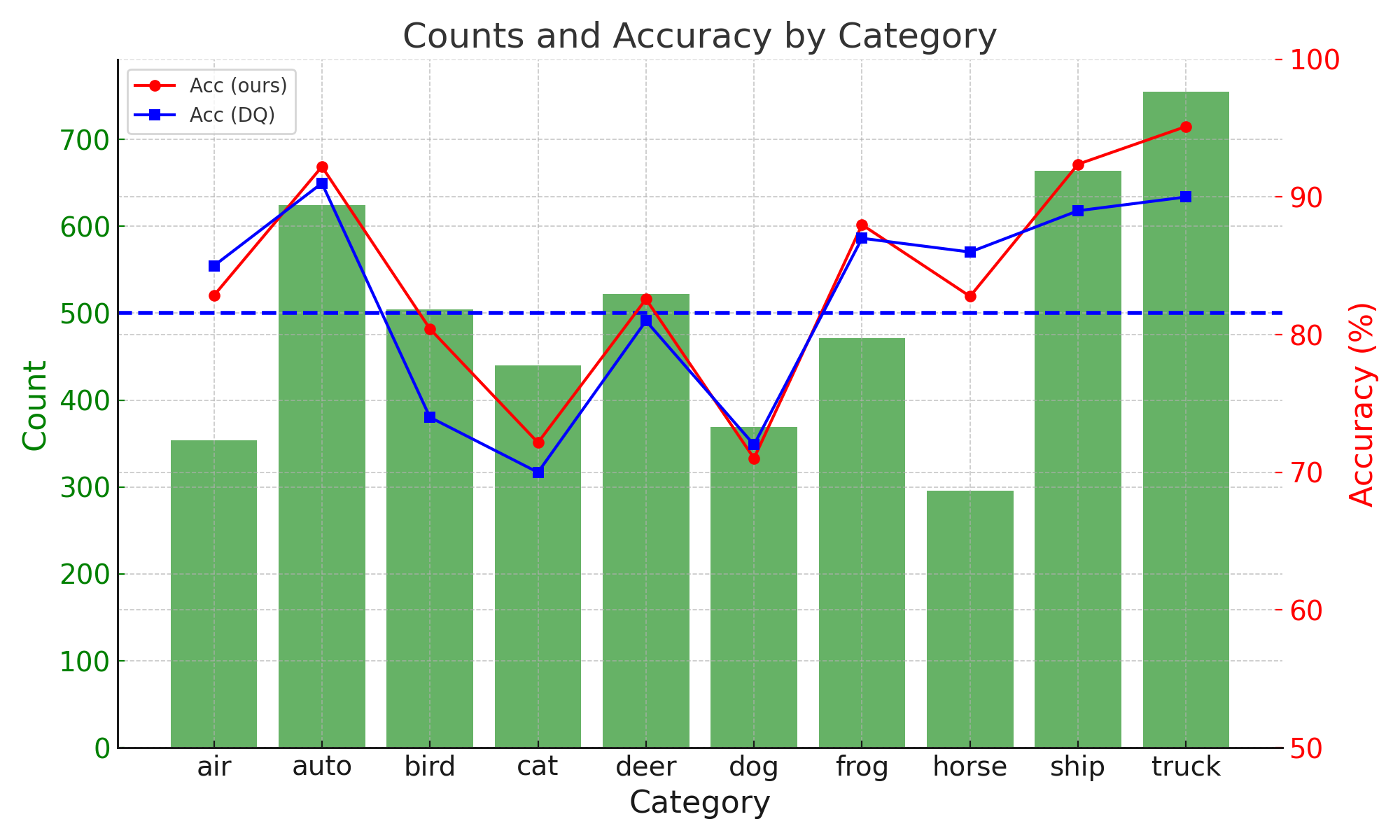}
    \caption{\textbf{Counts and accuracy by category comparison between DQ and DQAS.}}
    \label{fig:analyze}
    \vspace{-0.8cm}
\end{wrapfigure}

\subsection{Analysis}
In this section, we investigate the method effectiveness from the perspective of class-wise accuracy and the effectiveness perspective. 

\noindent \textbf{Analysis on class-wise sample counts and accuracy.}
We further analyze the accuracy for each class and the adaptive sample counts for each class on CIFAR-10. We put the class-wise counts and accuracies to Figure~\ref{fig:analyze}. For simplicity, ``airplane'' and ``automobile'' classes are labeled as ``air'' and ``auto'' in the figure.
This chart presents a comparative analysis of two different methods, our method DQAS and the baseline method (DQ), across various categories such as ``air'', ``auto'', ``bird'', ``cat'', etc, in CIFAR-10. We use the same AIPC for DQ and DQAS. The counts of samples for each class of our method are shown in green bars, and the IPC of DQ is shown as the blue dot line. The accuracy of each category of our method is shown as a red line, and the accuracy of each category of DQ is shown as a blue line.

There are some interesting observations from Figure~\ref{fig:analyze}. Firstly, it is notable that for certain categories such as ``airplane'', ``cat'', ``dog'', and ``frog'', despite a reduction in their counts, the accuracy of these classes still achieves competitive performance. This indicates that our method effectively compresses the dataset further by reducing the number of samples in classes that are not sensitive to sample size, thanks to active learning-based adaptive sampling.
Secondly, for other classes like ``automobile'', ``deer'', ``ship'', and ``truck'', an increase in the number of samples through our method appears to bolster model performance. This suggests that these categories are sensitive to the number of samples, requiring a larger dataset to address challenges associated with ``long tail'' data.
Thirdly, the ``bird'' class stands out as we barely change the counts for this class, yet observe a significant improvement in accuracy. This might be because the ``bird'' class benefits more from qualitative improvements in the dataset rather than the quantity of its samples. Enhancements such as more varied or representative samples, even in limited numbers, could be contributing to this substantial accuracy gain.

\begin{wraptable}{r}{0.4\textwidth}
  \vspace{-1cm}
  \centering
  \caption{\textbf{Computational cost.}}
  \begin{tabular}{c|c}
  \toprule
    Method & Condensing time \\
    \midrule
    DC~\cite{zhao2020datasetparameter_1} & 924.2 \\
    DREAM~\cite{liu2023dream} & 46.7 \\
    DQAS & \textbf{32.5} \\
    \bottomrule
  \end{tabular}
  \label{tab:cost}
  \vspace{-0.5cm}
\end{wraptable}

\noindent \textbf{Analysis of effectiveness of the method.}
We have compared the computational costs of our method with SOTA dataset condensation techniques, DC~\cite{zhao2020datasetparameter_1} and DREAM~\cite{liu2023dream}. We tested all three methods on an NVIDIA A5000 GPU for a fair comparison. We evaluated the GPU hours required for condensing the CIFAR-10 dataset at a 0.6 ratio. As detailed in Table~\ref{tab:cost}, our method has a significant advantage in efficiency. The class-wise dataset initialization enables our model to achieve a satisfactory performance with just a few steps of active learning, thereby significantly enhancing the computational efficiency of our method.

\vspace{-0.2cm}
\section{Conclusion}
\label{sec:conclusion}
\vspace{-0.2cm}
In this paper, we present an observation of stable and sensitive classes during dataset compression, which is prevalent across various dataset compression methods. Based on this observation, we propose a novel active learning based adaptive sampling to improve the performance of the dataset compression. Our adaptive sampling approach can be extended to many existing dataset compression methods. In addition, we introduce a new pipeline of dataset quantization, whose data feature is consistent across all stages of dataset quantization. 
We evaluate our approaches on CIFAR and Tiny ImageNet datasets to validate the effectiveness of our proposed method, and analyze the advantages of our method from both class-wise and efficiency perspectives. 

%
%
\bibliographystyle{splncs04}
\bibliography{main}
\end{document}